# Classifying Human Activities using Machine Learning and Deep Learning Techniques

## Sanku Satya Uday[1], Satti Thanuja Pavani[1], T.Jaya Lakshmi[1], Rohit Chivukula[2]

[1]Department of Computer Science and Engineering, SRM University, Andhra Pradesh, India.
[2]Department of Computer Science and Engineering, University of Huddersfield, United Kingdom.

*Abstract*

*Human Activity Recognition (HAR) describes the machine's ability to recognize human actions. Nowadays, most people on earth are health conscious, so people are more interested in tracking their daily activities using Smartphones or Smart Watches, which can help them manage their daily routines in a healthy way. With this objective, Kaggle has conducted a competition to classify 6 different human activities distinctly based on the inertial signals obtained from 30 volunteer's smartphones. The main challenge in HAR is to overcome the difficulties of separating human activities based on the given data such that no two activities overlap. In this experimentation, first, Data visualization is done on expert generated features with the help of t-distributed Stochastic Neighborhood Embedding(t-SNE) followed by applying various Machine Learning techniques like Logistic Regression, Linear SVC, Kernel SVM, Decision trees to better classify the 6 distinct human activities. Moreover, Deep Learning techniques like Long Short-Term Memory (LSTM), Bi-Directional LSTM, Recurrent Neural Network (RNN), and Gated Recurrent Unit (GRU) are trained using raw time series data. Finally, metrics like Accuracy, Confusion matrix, precision and recall are used to evaluate the performance of the Machine Learning and Deep Learning models. Experiment results proved that the Linear Support Vector Classifier in machine learning and Gated Recurrent Unit in Deep Learning provided better accuracy for human activity recognition compared to other classifiers.*

## 1. Introduction

In this modern era, the majority of individuals own a smartphone, which has become an integral part of human life. There are numerous sensors on these smartphones that assist in playing games, taking pictures, and for gathering other information. Apart from camera sensors, the accelerometer and gyroscope sensors, which record acceleration and angular velocity respectively, are the most significant sensors. By using these two sensors, University of California Irvine (UCI) created a dataset by experimenting with a group of 30 participants aged 19 to 48 years old. Each participant wore a smartphone (Samsung Galaxy S II) around their waist and did six activities (WALKING, WALKING UPSTAIRS, WALKING DOWNSTAIRS, SITTING, STANDING, LAYING). People at the institution captured Tri-axial linear acceleration and Tri-axial angular velocity at a constant rate of 50Hz using these embedded accelerometer and gyroscope sensors. The experiments were videotaped so that the data could be manually labelled. After collecting all the data in the form of frequency signals, the researchers at UCI, preprocessed the data by using some noise filters on the sensor signals and then those preprocessed signals were fabricated in the form of vectors which consists of 128 readings per sliding window of fixed width of 2.56 sec [1]. Here the task is to build multiclass classification models by using 128 dimensional fabricated vectors from signals of accelerometer and gyroscope sensors to map to one of those 6 types of daily human activities such as Walking, Walking-Upstairs, Walking-Downstairs, Sitting, Standing or Laying.

## 2. Problem Description

### 2.1. Problem Statement

Given accelerometer and gyroscope triaxial raw time series data of human activities in the form of vectors labelled with one of the 6 activity class labels (Walking, Walking-Upstairs, Walking-Downstairs, Sitting, Standing or Laying), the problem of human activities classification is to predict a class label from 1 to 6 different activities for an activity instance with missing label. As the prediction is not binary, but one among the 6 activities, this is a multi-class classification problem.

### 2.2. Dataset Description

UCI provided a dataset named UCI_HAR_Dataset in Kaggle platform, consisting of raw time series data as well as expert generated features. The raw time series data consists of tri-axial acceleration and tri-axial gyroscope data, where acceleration is divided into tri-axial body acceleration and tri-axial total acceleration [2]. There are a total 10,299 datapoints of 128 dimensions each where 70% of split is given to training phase and remaining 30% of data is given to testing phase. Whereas in expert generated features there are 561 features like tBodyAcc-mean(), tBodyAcc-std(), tBodyAcc-mean() and so on. Here 7,352 datapoints are used for training and 2,947 are used for testing phase.

## 2.3. Exploratory data analysis

For any problem the first thing to do is Exploratory Data Analysis. It is called exploratory because we don't know anything about the dataset we start with. So, performing Exploratory Data Analysis on a dataset aid in deeper understanding of the data. The distribution of instances over 6 classes is shown in Fig.1.

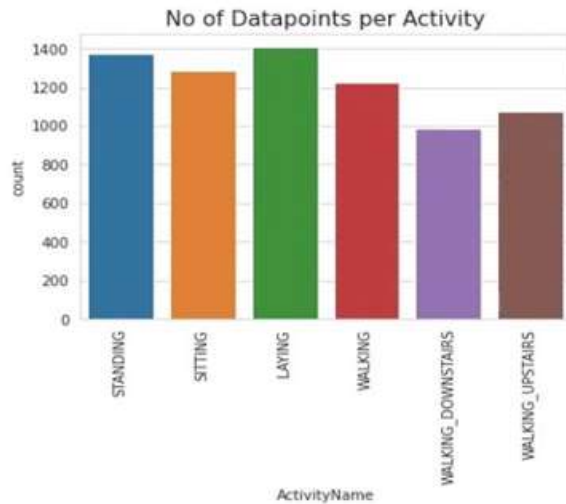

Figure 1. Distribution of instances over 6 classes

Univariate Analysis and t-SNE performed on the dataset, whose details are given below.

- Univariate Analysis: Univariate Analysis is a simple form of analysis that focuses on a single feature in order to extract useful insights from the data and identify meaningful patterns. Fig. 2 represents separation of stationary and moving activities with the help of Univariate Analysis on the feature "tBodyAccMagmean".

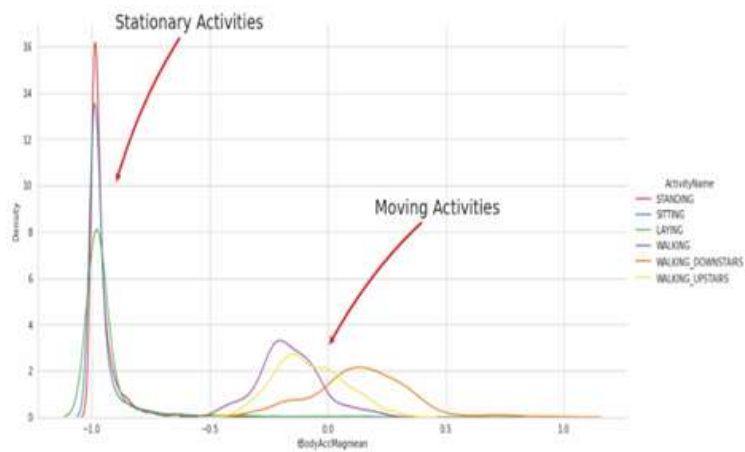

Figure 2. Univariate Analysis on feature "tBodyAccMagmean"

- T-SNE: T-distributed stochastic neighborhood embedding is a dimensionality reduction technique which is especially used for visualizing the data by minimizing the tendency to crowd points together in the center of the map [3]. It is a state of art technique which performs better than principal component analysis in especially reducing a data point from higher dimension to lower dimension. t-SNE preserves the datapoints within the cluster which helps to better classify the clusters in lower dimension. In this experimentation t-SNE is used to map higher dimensional space to lower dimensional space to better cluster all the six activities datapoints distinctly so that it can be better visualized. In t-distributed stochastic neighborhood embedding we have used expert generated features of 561 dimensional each to visualize the data in lower dimension. It is observed that there is an overlap between sitting and standing datapoints but remaining other 4 classes are clustered without any overlapping as shown in the Fig 3.

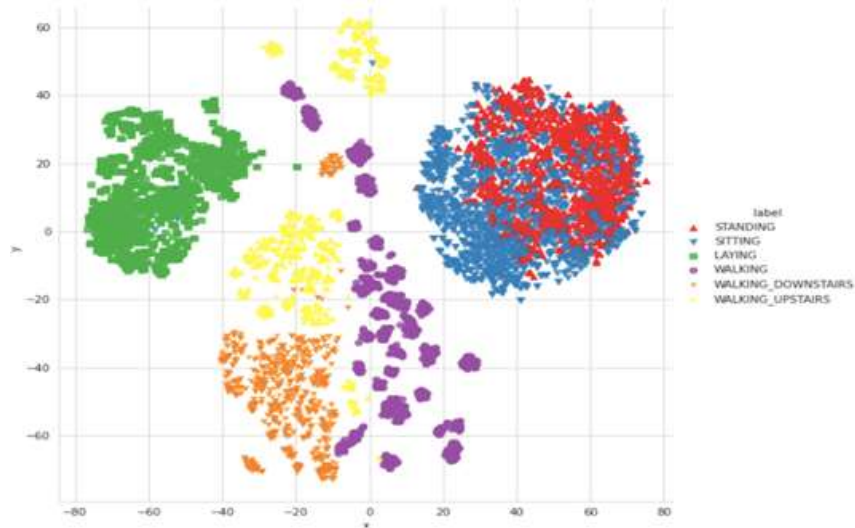

*Figure 3. T-SNE on 561 Expert generated Features*

## 3.  Related Literature

Human Activity Recognition is an ongoing research topic. It has various applications in different domains like fitness, medical field, gym, eldercare, surveillance and so on.

Saisakul Chernbumroong et al. performed an experiment using single wrist-worn sensor to detect 5 daily activities and states that the sensor's location is critical to the performance of activity recognition for human daily living in a free space environment. The use of a wrist-worn sensor might alleviate issues such as movement restriction, discomfort, and stigmatization [4].

Girija chetty et al. says that for processing these multiple sensor signals from smartphones for automatic and intelligent activity recognition, appropriate machine learning and data mining methods must be developed. Despite the fact that there are various machine learning approaches available it is unclear which algorithm will perform better for activity recognition on smartphones. Automatic activity recognition systems based on sophisticated processing of numerous sensors features on smart phones would be a huge help in the eHealth space, especially for remote activity monitoring and recognition in the aged care and disability care sectors [5].

Zameer Gulzar et al. finds that threshold-based algorithm is simpler and faster and is often used to detect the human activity. But the Machine Algorithm produces a reliable outcome [6]. Hamza Ali Imran et al. says that for problems like activity recognition Feature engineering should be prioritized instead of sophisticated deep learning architectures and says basic machine learning classifiers should be used [7].

Tahmina Zebin et al. focused their research on the use of deep learning approaches for human Activity Recognition. They used CNN to automate feature learning from multichannel time series data which gave the best results in classifying the human activities [8].  Seung Min Oh et al. says that to employ deep learning techniques, you must first learn using high-quality data sets. Adequate labelling is required for quality data, which takes time, effort, and cost of humans. Active learning is mostly used to reduce labelling time [9].

## 4.  Proposed Approach

The proposed approach is shown in Fig.4. The researchers generated 561 features using the raw time series data provided by UCI. Exploratory Data Analysis is performed on these features and found that "tBodyAccMagmean" separated the classes in the best way. Also, these experts generated features are used for training and testing the machine learning models while raw time series data is used for Deep learning models. Finally, Performance measures like Confusion matrix, Accuracy, Precision and Recall are computed to check which model performed better.

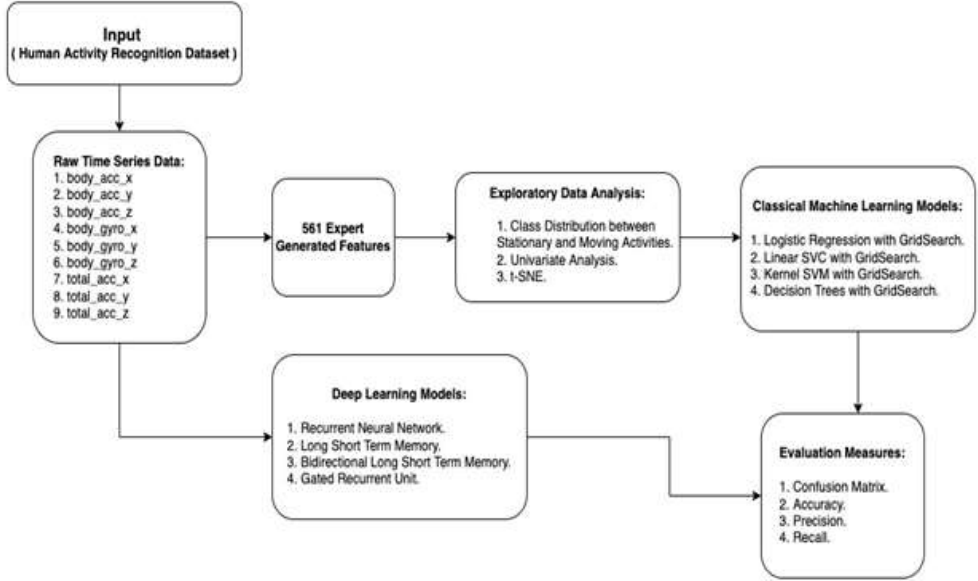

Figure 4. Proposed Approach

### 4.1. Machine Learning Models

- Logistic Regression: Logistic Regression is used for classification problems in machine learning which is used to estimate relationships between one or more independent variables. The objective of Logistic Regression is to find the best hyperplanes that separate 6 activities distinctly from each other in a linear fashion using one vs all approach [10]. Geometrically, the data points that lie above the hyperplane and are in the same direction as the weight vector belong to the positive class while the data points that lie in the opposite direction belong to the negative class. In order to find a hyperplane that best classifies the data points the parameters like weights(w) and bias(b) are required. Optimal weights for Logistic regression are computed using the equation 1.

$$W^* = arg_w \min \left[ \sum_{i=1}^{n} \log(1 + \exp(-y_i w^T x_i)) + \lambda w^T w \right] \tag{1}$$

After finding a best hyperplane with optimal weights and bias, to check whether the model is predicting accurately or not, a distance measure called signed distance is used for every single datapoint, but if we want to find the overall model performance the sum of signed distances fails when there are outliers. So, in order to overcome this problem, we use squashing technique which maps larger signed distance values to a value in between 0 to 1 by using sigmoidal activation function which gives good probabilistic interpretation. However, there might be a case, where overfitting might occur in logistic regression, which is due to outliers, leading weight values to +$\infty$ or -$\infty$ . To overcome this issue, regularize with grid search is used to control weight values with the help of hyperparameter in such a way that the model does not overfit. Here Logistic loss is used as an approximation for 0-1 loss to make loss function continuous such that it is differentiable [11].

- Linear SVM: Support Vector Machines are used for classifying linear and nonlinear data, which is used in many areas of application, especially in dealing with high dimensional data [12]. In this problem the main goal of SVC is to find margin maximizing hyperplanes which better classifies six distinct human activities. If the margin distance is high, the chance that the points misclassify decreases. To construct an SVC, first construct a convex hull for similar data points and then find the shortest lines connecting these hulls and then bisect these connecting lines by using a plane. These bisecting planes are called as margin maximizers which better classifies human activities. Here hinge loss is used as an approximation for 0-1 Loss [13]. The weights and bias for margin maximizing hyperplanes are computed using the equation 2.

$$(w^*, b^*) = argmin_{w,b} \frac{\|w\|}{2} + c.\frac{1}{n}\sum_{i=1}^{n} \xi_i \qquad (2)$$

- RBF Kernel SVM: The kernel SVM is an extension of soft margin SVM where kernel trick is additionally operated. Here in this research, radial basis function is used as a kernel which helps in handling nonlinear data. Kernelization takes data which is in d dimensional space and performs feature transformation internally in such a way that nonlinear data can be linearly separable in transformed space d' [14]. For doing such transformation, predominantly two parameters named c in soft margin SVM and sigma in RBF are required, which can be optimally found by using grid search. The overall computation of margin maximizing hyperplanes by using kernel trick is computed with the help of equation 3.

$$\max_{\alpha_i} \sum_{i=1}^{n} \sum_{j=1}^{n} \alpha_i . \alpha_j . y_i . y_j . rbf\_kernel(x_i, x_j) \qquad (3)$$

- Decision Trees: Decision trees are used for both classification and regression problems in machine learning.This classifier is very similar to if .. else logical conditions in programming. At leaf nodes of the decision tree, class labels are decided while all non-leaf nodes are involved in the decision-making process. Corresponding to every decision there exists a hyperplane which is axis parallel. Therefore, the decision tree classifies all the distinct human activities axis parallelly using hyperplanes [15].

## 4.2. Deep Learning Models

- RNN: Recurrent Neural Networks are mainly used to process sequential data. Here neural networks are termed as recurrent because the information cycles through a loop with time. In RNN, 3 types of weights need to be initialized based on the activation unit. If the activation unit is tanh or sigmoidal, xavier or glorat initialization is used or if the activation unit is relu, "He" initialization is used. RNN can only take care about short term dependencies, but not long-term dependencies, this will cause problems in both forward propagation and backward propagation because in real world there might be a requirement of long-term dependencies where the input, which is learned long back ago, should be retained to the current state [16].

- LSTM: Long Short-Term Memory network models are advancement of recurrent neural networks that can learn and remember extended sequences of input data. They're designed to work with data that's made up of extended sequences of data, ranging from 200 to 400-time steps. LSTM's are good fit for this problem. Multiple parallel sequences of input data, such as each axis of the accelerometer and gyroscope data, are supported by this model. The model learns how to extract features from observation sequences and how to map internal features to distinct activity types [16]. The advantage of using LSTMs for sequence classification is that they can learn directly from raw time series data, eliminating the need for domain knowledge to manually construct input features.

The model is able to learn an internal representation of the time series data and is also useful to retain the cell state. The main feature of LSTM is short circuit i.e, from $C_{t-1}$ (previous cell state output) to $C_t$ (Current cell state output). The output from previous cell state will not be disturbed if short-circuited ($C_t = C_{t-1}$). If the LSTM needs to be short circuited the forget gate output should be list of one's and later, point wise addition(+) from the input gate should be list of zeroes. From the output of the forget gate we can decide how much of the previous cell state to be remembered or how much should this cell state forget. When coming to the input gate, the output of this gate tells how much new information should be added. Finally, the output of the cell state is released through the output gate.

- Bidirectional LSTM: Bidirectional Long Short Term Memory is an extension of LSTM which can be used in learning both forward and backward sequence patterns of the sequential input data. In this model later inputs can also impact the previous outputs. Here the training is done on two LSTM instead of one on input sequence. The first model learns the sequence of input provided whereas the second model learns the reverse of that sequence.

This helps the model to have additional context such that the model can also improve its performance [17].

- GRU: Gated Recurrent units are a gating mechanism in recurrent neural networks. GRU is the same as LSTM but with only two gates named reset and update gate respectively. In this there is no need for a cell state because everything is captured in output itself. Due to less complicated architecture than LSTM, the GRU performs faster computation on partial derivatives which results in less time complexity [18]. GRU has less parameters (4230) when compared to LSTM(5574). The 9 raw time series data namely body_acc_x, body_acc_y, body_acc_z, body_gyro_x, body_gyro_y, body_gyro_z, total_acc_x, total_acc_y, $total\_acc\_z$ are given as input to 32 GRU 's(Gated Recurrent unit) in the form of 128 dimensional vectors. After giving input to GRU's the dropout of 0.5 is added because the no.of parameters(4230) are closer to no.of datapoints(7352) in the training dataset , So it is very trivial to overfit the model. After generating the output from GRU's every output is connected to the neurons present in the dense layer. The dense layer ensures that all the outputs from GRU's are fully connected. Now, from the dense layer, by using the function softmax it predicts one of the distinct human activities according to the generated output data.

## 5. Results

In this experimentation, as machine learning models are unable to fabricate features directly from raw time series data, the features that are generated by experts from the raw data are given as input to train the machine learning models. However, deep learning algorithms use the raw time series data. Accuracy of various human activities recognised using machine learning algorithms are given in Table.1

*Table 1. Accuracy of Machine learning algorithms using expert generated features*

| Algorithm | LAYING | SITTING | STANDING | WALKING | WALKING DOWNSTAIRS | WALKING UPSTAIRS |
|---|---|---|---|---|---|---|
| Logistic Regression | 100% | 88% | 97% | 99% | 96% | 95% |
| Linear SVM | 100% | 88% | 98% | 100% | 98% | 96% |
| RBF Kernel SVM | 100% | 90% | 98% | 99% | 95% | 96% |
| Decision Trees | 100% | 75% | 89% | 95% | 84% | 77% |

Three of the Machine Learning models better classified all the six basic human activity labels but with a slight confusion between sitting and standing data points. All in all Linear Support Vector Classifier performed exceptionally well on expert generated features with 96.7% accuracy, while Decision Tree model performed comparatively less with an accuracy of 87%. The confusion matrix of Linear SVM is given in Fig. 5.

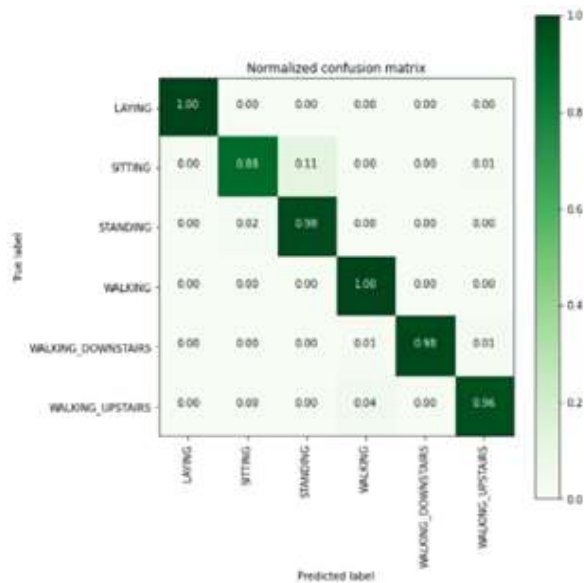

*Figure 5. Confusion Matrix of Linear SVM*

On the other hand, Deep learning model's used in this research are able to generate features on their own using the raw time series data. The accuracy is tabulated in Table. 2 and the confusion matrices LSTM and GRU are given in Fig. 6 and Fig.7. Here Recurrent Neural Networks used 1,542 parameters from raw time series data for training and performed 77.64% accurately, while the Gated Recurrent Unit model used 4,230 parameters and achieved an accuracy of 92.60% which performed better than all other deep learning models.

*Table 2. Accuracy of Deep Learning algorithms using raw time series data*

| Algorithm | LAYING | SITTING | STANDING | WALKING | WALKING DOWNSTAIRS | WALKING UPSTAIRS |
|---|---|---|---|---|---|---|
| RNN | 99% | 88% | 55% | 45% | 94% | 89% |
| LSTM | 95% | 77% | 89% | 95% | 98% | 97% |
| Bidirectional LSTM | 100% | 74% | 82% | 86% | 98% | 99% |
| GRU | 97% | 78% | 90% | 99% | 94% | 97% |

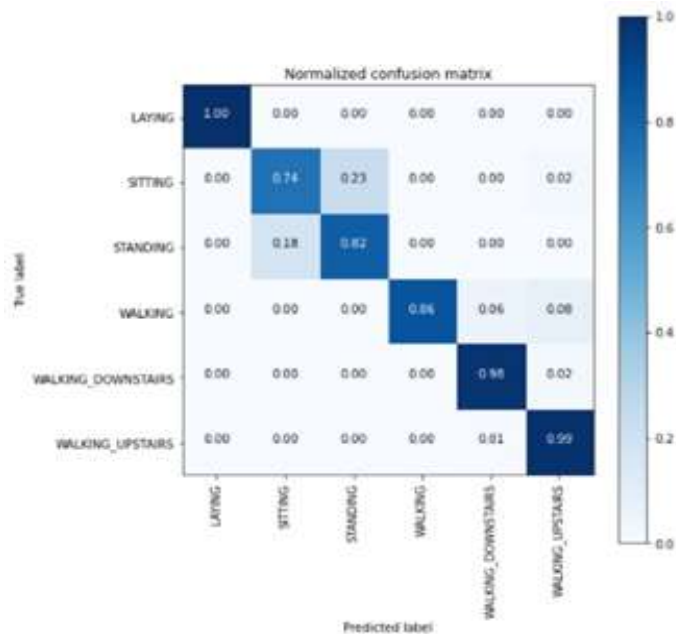

*Figure 6 Confusion Matrix of LSTM*

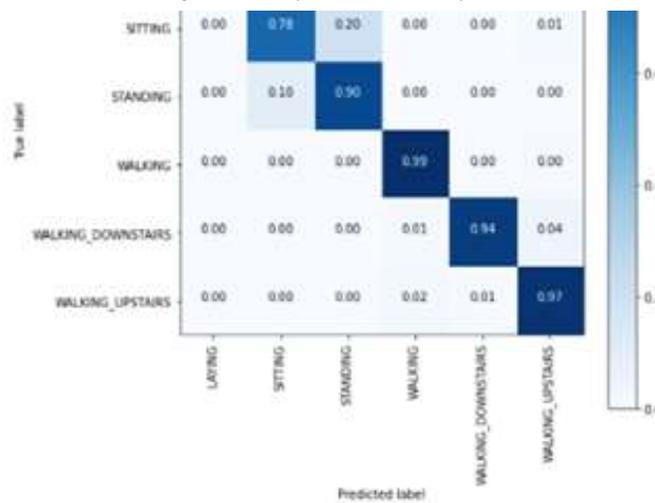

*Figure 5. Confusion Matrix of GRU*

## 6. Conclusion and Future Scope

In classifying the 6 distinct human activities, the machine learning models performed better than deep learning models due to the fact that machine learning models are provided with expert generated features which gives better understanding to the model. However, Deep Learning models showed no less performance even though they generated their own features using the raw time series data. In case of Machine learning Models, Linear Support Vector Classifiers achieved an highest accuracy of 96.7\% , whereas in case of Deep learning Models, Gated Recurrent Unit achieved an highest accuracy of 92.60\%. Even Though these models performed their best, both Deep Learning and Machine Learning Models faced the same confusion between sitting and standing activities because of the reason that these two activities are stationary. So, In future our goal is to train the models in such a way that it avoids confusion between these stationary activities such that these findings could be utilised to create smart watches and other devices that track a user's activities and alert him or her of the daily activity record.